\pdfoutput=1

\documentclass[11pt]{article}

\usepackage[final]{acl}

\usepackage{times}
\usepackage{latexsym}
\usepackage{booktabs}
\usepackage{adjustbox}
\usepackage{pifont}
\usepackage[T1]{fontenc}

\usepackage[utf8]{inputenc}

\usepackage{microtype}

\usepackage{inconsolata}

\usepackage{graphicx}
\usepackage{subfiles} 

\usepackage{enumitem}
\usepackage{cleveref}
\usepackage{xspace}
\makeatletter
\DeclareRobustCommand\onedot{\futurelet\@let@token\@onedot}
\def\@onedot{\ifx\@let@token.\else.\null\fi\xspace}
\def\eg{\emph{e.g}\onedot} 
\def\ie{\emph{i.e}\onedot}

\makeatother
\usepackage{soul}

\title{Document Intelligence in the Era of Large Language Models: A Survey}

\newcommand*{\affaddr}[1]{#1} 
\newcommand*{\affmark}[1][*]{\textsuperscript{#1}}
\newcommand*{\email}[1]{\textrm{#1}}

\author{
  Weishi Wang\affmark[1]~~
  Hengchang Hu\affmark[1]~~
  Zhijie Zhang\affmark[1]~~ \\
  \textbf{Zhaochen Li\affmark[1]}~~ 
  \textbf{Hongxin Shao\affmark[1]}~~
  \textbf{Daniel Dahlmeier\affmark[1]}~~ \\
\affaddr{\affmark[1]SAP, Singapore} \\
\email{\small{\{weishi.wang, hengchang.hu, zhijie.zhang, zhaochen.li, hongxin.shao, d.dahlmeier\}@sap.com}}\\
}

\begin{document}
\maketitle
\begin{abstract}
Document AI (DAI) has 
emerged as a vital application area, and is significantly transformed by the advent of large language models (LLMs). While earlier approaches relied on encoder-decoder architectures, decoder-only LLMs have revolutionized DAI, bringing remarkable advancements in understanding and generation. This survey provides a comprehensive overview of DAI’s evolution, highlighting current research attempts and future prospects of LLMs in this field. We explore key advancements and challenges in multimodal, multilingual, and retrieval-augmented DAI, while also suggesting future research directions, including agent-based approaches and document-specific foundation models. This paper aims to provide a structured analysis of the state-of-the-art in DAI and its implications for both academic and practical applications. 
\end{abstract}

\section{Introduction}

Documents serve as diverse formats for conveying information, playing crucial roles in both research and industry~\cite{stage2003research}. Document AI (DAI) leverages natural language processing and computer vision techniques to automate document-related tasks in two categories: understanding~\cite{DBLP:journals/corr/abs-2206-01062, DBLP:conf/wacv/MathewKJ21} and generation~\cite{wiseman2017challenges,DBLP:conf/emnlp/ZhangYY18}. However, traditional rule-based~\cite{DBLP:conf/icpr/BourgeoisBE92} and learning-based~\cite{DBLP:journals/pami/MarinaiGS05} approaches often prove cost-prohibitive and time-consuming, motivating the shift towards language models for more efficient and scalable solutions.

Among these models, \textit{encoder-only} models \cite{DBLP:conf/aaai/HongKJHNP22,DBLP:conf/mm/HuangL0LW22,DBLP:conf/cvpr/LuoCZY23} are widely used to capture nuanced, document-specific information, while \textit{encoder-decoder} architectures excel in generation tasks involving variable-length and multimodal inputs and outputs~\cite{DBLP:conf/aaai/TanakaNY21,DBLP:conf/cvpr/TangYWFLZ0ZB23}, though they remain constrained by their training data. 
Recent advancements in \textit{decoder-only} Large Language Models (LLMs) have spurred the development of document-focused LLMs trained on diverse multimodal datasets, such as image segmentation~\cite{DBLP:conf/emnlp/HuXYYZZZJHZ24}, layout~\cite{DBLP:conf/acl/0005RSMBKPNL24,DBLP:conf/cvpr/LuoSZZYY24}, and geometric information~\cite{DBLP:conf/icdar/LamottWUSKO24}, demonstrating promising performance in both understanding and generating document content. Despite recent successes, LLMs struggle to interpret documents accurately~\cite{DBLP:journals/tacl/LiuLHPBPL24}, often relying on Optical Character Recognition (OCR) engines or neglecting the rich textual information within documents. Such limitations hinder the learning of a unified representation across multiple document-specific modalities. Moreover, there remain pervasive challenges in multilingualism and linguistic diversity~\cite{DBLP:conf/acl/JoshiSBBC20}, as data scarcity and the lack of high-quality training data restrict the performance of multilingual LLMs. While the effectiveness of cross-lingual generalization has been explored for translation-equivalent tasks~\cite{DBLP:conf/emnlp/ZhangLHSK23}, DAI often requires complex contextual information, including document structure and language-specific knowledge~\cite{DBLP:journals/corr/abs-2410-12628,DBLP:conf/emnlp/XingCGSYBZY24}, which can further limit LLM performance in multilingual document processing.

The development and application of LLMs in DAI have emerged as a key research area. Multilingual and multimodal capabilities are essential for enabling these models to effectively handle diverse, real-world scenarios~\cite{DBLP:journals/corr/abs-2404-00929,DBLP:journals/corr/abs-2402-03173}. Recent efforts have focused on integrating multilingualism and multimodality to enhance unified document representation learning, thereby improving the versatility and robustness of LLMs~\cite{DBLP:conf/emnlp/HuXYYZZZJHZ24}. Given the rapid evolution of LLM-based DAI, there is an increasing need for a comprehensive overview of their capabilities in document processing. To address this need, we survey recent advancements in state-of-the-art (SoTA) multilingual and multimodal LLMs, highlighting their progress in DAI, the challenges they encounter, and future prospects.

In this in-depth survey paper, we systematically categorize relevant research on LLM-based DAI into five key tasks. Our contributions include:
\begin{itemize}[leftmargin=*,topsep=2pt,itemsep=2pt,parsep=0pt]
\item We introduce the detailed definition and trace the evolution of document intelligence, along with an overview of relevant benchmarks in \Cref{sec_2}.
\item We comprehensively review recent advancements in integrating multimodal capabilities in  LLMs, emphasizing their effectiveness on diverse DAI applications in \Cref{sec_4:multimodal}. 
\item We systematically analyze the advancements in the multilingual capabilities (\Cref{sec_3:multilingual}) and retrieval-augmented paradigm (\Cref{rag}), exploring their impact on improving context-aware document intelligence. 
\item We summarize key technical areas addressing current challenges in developing reliable document-specific foundation models, outlining actionable prospects for future developments in DAI.
\end{itemize}
\section{Task Definitions and Benchmarks}
\label{sec_2}
\noindent\textbf{Task Definitions}~~~
Research in DAI has gained significant attention and can be broadly categorized into document-related \emph{understanding} and \emph{generation} tasks. We define understanding tasks in DAI as those aimed at extracting and analyzing information from existing documents, while generation tasks in DAI involve creating new content and responses based on given documents and instructions. In essence, understanding focuses on encoding information within documents, while generation emphasizes decoding and producing content and responses. Notably, these two categories often overlap. Multilingual and multi-document pose additional challenges to DAI. However, given their frequent occurrence in real-world applications, they are crucial for real-world DAI~\cite{DBLP:journals/corr/abs-2104-08836, DBLP:journals/corr/abs-2409-03420}. Relevant research works in the LLM era will be discussed in Section \ref{sec_4:multimodal}-\ref{rag}, while early work and evolution are provided in \Cref{appx:evolution}.

For the understanding tasks, Key Information Extraction (KIE) involves extracting essential information from unstructured or semi-structured documents~\cite{DBLP:journals/corr/abs-2403-04080}. This process typically comprises two components: OCR for converting images of text into machine-readable format, and named entity recognition (NER) for classifying entities into predefined categories~\cite{DBLP:journals/tkde/LiSHL22}. Document Layout Analysis (DLA) focuses on analyzing the spatial arrangement of documents to understand their structure and layout, which is crucial for various downstream applications~\cite{DBLP:journals/csur/BinMakhashenM20}. For Document-level tasks, Document Sentiment Analysis (DSA) determines the sentiment expressed in documents, providing insights into the emotional tone of the context, while Document Classification (DC) categorizes documents based on their content.

In the generation category, Document Summarization (DS) aims to create concise summaries of documents~\cite{Yao2017RecentAI, Ma2020MultidocumentSV}. Document Content Generation (DCG) involves generating new document content based on existing materials, such as generating textual continuations or creating figures and tables derived from the provided documents~\cite{DBLP:journals/corr/abs-2406-08354}.
Question Answering (QA) focuses on generating accurate natural language responses to a given question, based on the context of documents~\cite{Zhu2021RetrievingAR, DBLP:conf/wacv/MathewKJ21}. However, QA can involve both understanding and generation tasks, \ie, extractive and generative QA. Given the focus LLMs, we classify it primarily as a generation task.

\noindent\textbf{Benchmarks and Datasets}~~~
Numerous comprehensive benchmarks have been established to evaluate DAI. A summary of these benchmarks, detailing languages, document counts, modalities, tasks, and open-source availability, is presented in~\Cref{table:benchmark}.

\begin{table*}
\begin{adjustbox}{width=1\textwidth}
\begin{tabular}{r|rrrr |rrr |rrrrrrr}
\toprule
& ~ & ~ & ~ & ~& \multicolumn{3}{|c}{Modality} & \multicolumn{7}{|c}{Task} \\
\cmidrule(r){6-8} \cmidrule(l){9-15}
Dataset & Language &\#Languages &\#Documents & Open-Source & Text & Visual & Layout & KIE & DLA & DSA & DC & DS & DCG & QA \\
\midrule
SROIE~\cite{DBLP:conf/icdar/HuangCHBKLJ19}  & EN & 1 & 973 & \ding{51} & \ding{51} & \ding{51} & \ding{51} & \ding{51} & \ding{51} & ~ & ~ & ~ & ~ & ~ \\ 
FUNSD~\cite{DBLP:conf/icdar/JaumeET19} & EN & 1 & 199 & \ding{51} & \ding{51} & \ding{51} & \ding{51} & \ding{51} & ~ & ~ & ~ & ~ & ~ & ~ \\ 
CORD~\cite{park2019cord} & EN & 1 & 1,000 & \ding{51} & \ding{51} & \ding{51} & \ding{51} & \ding{51} & \ding{51} & ~ & ~ & ~ & ~ & ~ \\ 
CUTIE~\cite{DBLP:journals/corr/abs-1903-12363} & ES & 1 & 4,484 & \ding{56} & \ding{51} & \ding{51} & \ding{51} & \ding{51} & ~ & ~ & ~ & ~ & ~ & ~ \\ 
Kleister NDA~\cite{DBLP:conf/icdar/StanislawekGWLK21} & EN & 1 & 540 & \ding{51} & \ding{51} & \ding{51} & \ding{51} & \ding{51} & ~ & ~ & ~ & ~ & ~ & ~ \\ 
Kleister Charity~\cite{DBLP:conf/icdar/StanislawekGWLK21} & EN & 1 & 2,788 & \ding{51} & \ding{51} & \ding{51} & \ding{51} & \ding{51} & ~ & ~ & ~ & ~ & ~ & ~ \\ 
DWIE~\cite{DBLP:journals/ipm/ZaporojetsDDD21} & EN & 1 & 802 & \ding{51} & \ding{51} & \ding{51} & \ding{51} & \ding{51} & \ding{51} & ~ & ~ & ~ & ~ & ~ \\ 
WildReceipt~\cite{DBLP:journals/corr/abs-2103-14470} & EN & 1 & 1,768 & \ding{51} & \ding{51} & \ding{51} & \ding{51} & \ding{51} & \ding{51} & ~ & ~ & ~ & ~ & ~ \\ 
MC-OCR~\cite{DBLP:conf/rivf/VuBNNV21} & Multi & 2 & 2,436 & \ding{51} & \ding{51} & \ding{51} & \ding{51} & \ding{51} & ~ & ~ & ~ & ~ & ~ & ~ \\ 
EPHOIE~\cite{DBLP:conf/aaai/WangLJT0ZWWC21} & CN & 1 & 1,494 & \ding{51} & \ding{51} & \ding{51} & \ding{51} & \ding{51} & ~ & ~ & ~ & ~ & ~ & ~ \\ 
XFUND~\cite{DBLP:conf/acl/XuL0WLFZW22} & Multi & 7 & 1,393 & \ding{51} & \ding{51} & \ding{51} & \ding{51} & \ding{51} & ~ & ~ & ~ & ~ & ~ & ~ \\ 
VRDU Registration~\cite{DBLP:conf/kdd/WangZWLT23} & EN & 1 & 1,915 & \ding{51} & \ding{51} & \ding{51} & \ding{51} & \ding{51} & ~ & ~ & ~ & ~ & ~ & ~ \\ 
VRDU Ad-buy~\cite{DBLP:conf/kdd/WangZWLT23} & EN & 1 & 641 & \ding{51} & \ding{51} & \ding{51} & \ding{51} & \ding{51} & ~ & ~ & ~ & ~ & ~ & ~ \\ 
DocILE~\cite{DBLP:conf/icdar/SimsaSUPHKSMDCK23} & EN & 1 & 6,680 & \ding{51} & \ding{51} & \ding{51} & \ding{51} & \ding{51} & ~ & ~ & ~ & ~ & ~ & ~ \\ 
POIE~\cite{DBLP:conf/icdar/KuangHLYJRB23} & EN & 1 & 3,000 & \ding{51} & \ding{51} & \ding{51} & \ding{51} & \ding{51} & ~ & ~ & ~ & ~ & ~ & ~ \\ 
AMuRD~\cite{DBLP:journals/corr/abs-2309-09800} & Multi & 2 & 47,720 & \ding{51} & \ding{51} & ~ & ~ & \ding{51} & ~ & ~ & \ding{51} & ~ & ~ & ~ \\ 
UIT-MLReceipts~\cite{nguyen2024uit} & Multi & 2 & 2,147 & \ding{56} & \ding{51} & \ding{51} & \ding{51} & \ding{51} & ~ & ~ & ~ & ~ & ~ & ~ \\ 
CORU~\cite{DBLP:journals/corr/abs-2406-04493} & Multi & 2 & 20,000 & \ding{51} & \ding{51} & \ding{51} & \ding{51} & \ding{51} & \ding{51} & ~ & ~ & ~ & ~ & ~ \\ 
DocRED~\cite{DBLP:conf/acl/YaoYLHLLLHZS19} & EN & 1 & 5,053 & \ding{51} & \ding{51} & \ding{51} & \ding{51} & ~ & \ding{51} & ~ & ~ & ~ & ~ & ~ \\ 
PubLayNet~\cite{DBLP:conf/icdar/ZhongTJ19} & N/A & N/A & 358,353 & \ding{51} & \ding{51} & \ding{51} & \ding{51} & ~ & \ding{51} & ~ & ~ & ~ & ~ & ~ \\ 
TableBank~\cite{DBLP:conf/lrec/LiCHWZL20} & Multi & N/A & 18,000 & \ding{51} & \ding{51} & \ding{51} & \ding{51} & ~ & \ding{51} & ~ & ~ & ~ & ~ & ~ \\ 
DocBank~\cite{DBLP:conf/coling/LiXCHWLZ20} & EN & 1 & 500,000 & \ding{51} & \ding{51} & \ding{51} & \ding{51} & ~ & \ding{51} & ~ & ~ & ~ & ~ & ~ \\ 
ReadingBank~\cite{DBLP:conf/emnlp/WangX0SW21} & EN & 1 & 500,000 & \ding{51} & \ding{51} & \ding{51} & \ding{51} & ~ & \ding{51} & ~ & ~ & ~ & ~ & ~ \\ 
arXivdocs-target~\cite{DBLP:conf/aaai/RauschMB0F21} & EN & 1 & 362 & \ding{51} & \ding{51} & \ding{51} & \ding{51} & ~ & \ding{51} & ~ & ~ & ~ & ~ & ~ \\
DocLayNet~\cite{DBLP:journals/corr/abs-2206-01062} & Multi & 4 & 80,863 & \ding{51} & \ding{51} & \ding{51} & \ding{51} & ~ & \ding{51} & ~ & ~ & ~ & ~ & ~ \\ 
HRDoc~\cite{DBLP:conf/aaai/MaDHZZZL23} & EN & 1 & 2,500 & \ding{51} & \ding{51} & \ding{51} & \ding{51} & ~ & \ding{51} & ~ & ~ & ~ & ~ & ~ \\ 
E-Periodica~\cite{DBLP:conf/icdm/RauschRG0F23} & Multi & 4 & 542 & \ding{51} & \ding{51} & \ding{51} & \ding{51} & ~ & \ding{51} & ~ & ~ & ~ & ~ & ~ \\ 
DocHieNet~\cite{DBLP:conf/emnlp/XingCGSYBZY24} & Multi & 2 & 1,673 & \ding{51} & \ding{51} & \ding{51} & \ding{51} & ~ & \ding{51} & ~ & ~ & ~ & ~ & ~ \\ 

Amazon Sentiment Polarity~\cite{DBLP:conf/acl/PrettenhoferS10} & Multi & 4 & 6,000 & \ding{51} & \ding{51} & ~ & ~ & ~ & ~ & \ding{51} & ~ & ~ & ~ & ~ \\ 
Sinhala-English-Code-Switched-Dataset~\cite{DBLP:journals/kais/RathnayakeSRR22} & Multi & 2 & 10,000 & \ding{51} & \ding{51} & ~ & ~ & ~ & ~ & \ding{51} & ~ & ~ & ~ & ~  \\ 
AfriSenti~\cite{DBLP:conf/emnlp/MuhammadAAOAYAB23} & Multi & 14 & 112,506 & \ding{51} & \ding{51} & ~ & ~ & ~ & ~ & \ding{51} & ~ & ~ & ~ & ~  \\ 
Tobacco-3482~\cite{DBLP:journals/prl/KumarYD14} & EN & 1 & 3,482 & \ding{51} & ~ & \ding{51} & ~ & ~ & ~ & ~ & \ding{51} & ~ & ~ & ~ \\ 
RVL-CDIP~\cite{DBLP:conf/icdar/HarleyUD15} & EN & 1 & 400,000 & \ding{51} & ~ & \ding{51} & ~ & ~ & ~ & ~ & \ding{51} & ~ & ~ & ~ \\ 
MLDoc~\cite{DBLP:conf/lrec/SchwenkL18} & Multi & 8 & 6,000 & \ding{51} & \ding{51} & ~ & ~ & ~ & ~ & ~ & \ding{51} & ~ & ~ & ~ \\ 
MultiEURLEX~\cite{DBLP:conf/emnlp/ChalkidisFA21} & Multi & 23 & 65,000 & \ding{51} & \ding{51} & ~ & ~ & ~ & ~ & ~ & \ding{51} & ~ & ~ & ~ \\ 
En2ZhSum~\cite{DBLP:conf/emnlp/ZhuWWZZWZ19} & Multi & 2 & 370,687 & \ding{51} & \ding{51} & ~ & ~ & ~ & ~ & ~ & ~ & \ding{51} & ~ & ~ \\ 
Zh2EnSum~\cite{DBLP:conf/emnlp/ZhuWWZZWZ19} & Multi & 2 & 1,699,713 & \ding{51} & \ding{51} & ~ & ~ & ~ & ~ & ~ & ~ & \ding{51} & ~ & ~ \\ 
MLSUM~\cite{DBLP:conf/emnlp/ScialomDLPS20} & Multi & 5 & 1,571,041 & \ding{51} & \ding{51} & ~ & ~ & ~ & ~ & ~ & ~ & \ding{51} & ~ & ~ \\ 
Wikilingua~\cite{DBLP:conf/emnlp/LadhakDCM20} & Multi & 18 & 770,000 & \ding{51} & \ding{51} & ~ & ~ & ~ & ~ & ~ & ~ & \ding{51} & ~ & ~ \\ 
XL-Sum~\cite{DBLP:conf/acl/HasanBIMLKRS21} & Multi & 44 & 1,005,292 & \ding{51} & \ding{51} & ~ & ~ & ~ & ~ & ~ & ~ & \ding{51} & ~ & ~ \\ 
MassiveSumm~\cite{DBLP:conf/emnlp/VarabS21} & Multi & 92 & 31,940,180 & \ding{51} & \ding{51} & ~ & ~ & ~ & ~ & ~ & ~ & \ding{51} & ~ & ~ \\ 
CrossSum~\cite{DBLP:conf/acl/BhattacharjeeHA23} & Multi & 45 & 1,678,466 & \ding{51} & \ding{51} & ~ & ~ & ~ & ~ & ~ & ~ & \ding{51} & ~ & ~ \\ 
CroCoSum~\cite{DBLP:conf/coling/ZhangE24} & Multi & 2 & 42,000 & \ding{51} & \ding{51} & ~ & ~ & ~ & ~ & ~ & ~ & \ding{51} & ~ & ~ \\ 
GlobeSumm~\cite{DBLP:conf/emnlp/YeFFM0XYL024} & Multi & 26 & 4,687 & \ding{51} & \ding{51} & ~ & ~ & ~ & ~ & ~ & ~ & \ding{51} & ~ & ~ \\ 
Crello~\cite{DBLP:conf/iccv/Yamaguchi21} & N/A & N/A & 23,322 & \ding{51} & ~ & \ding{51} & \ding{51} & ~ & ~ & ~ & ~ & ~ & \ding{51} & ~ \\
PubGenNet~\cite{DBLP:journals/corr/abs-2406-08354} & N/A & N/A & 346,948 & \ding{56} & \ding{51} & \ding{51} & \ding{51} & ~ & ~ & ~ & ~ & ~ & \ding{51} & ~ \\ 

TriviaQA~\cite{DBLP:conf/acl/JoshiCWZ17} & EN & 1 & 662,659 & \ding{51} & \ding{51} & ~ & ~ & ~ & ~ & ~ & ~ & ~ & ~ & \ding{51} \\ 
HotpotQA~\cite{DBLP:conf/emnlp/Yang0ZBCSM18} & EN & 1 & 112,779 & \ding{51} & \ding{51} & ~ & ~ & ~ & ~ & ~ & ~ & ~ & ~ & \ding{51} \\ 
NarrativeQA~\cite{DBLP:journals/tacl/KociskySBDHMG18} & EN & 1 & 1,572 & \ding{51} & \ding{51} & ~ & ~ & ~ & ~ & ~ & ~ & ~ & ~ & \ding{51} \\ 
TextVQA~\cite{singh2019towards} & EN & 1 & 28,408 & \ding{51} & \ding{51} & \ding{51} & \ding{51} & ~ & ~ & ~ & ~ & ~ & ~ & \ding{51} \\ 
OCR-VQA~\cite{DBLP:conf/icdar/0001SSC19} & EN & 1 & 207,572 & \ding{51} & \ding{51} & \ding{51} & ~ & ~ & ~ & ~ & ~ & ~ & ~ & \ding{51} \\ 
Natural Questions~\cite{DBLP:journals/tacl/KwiatkowskiPRCP19} & EN & 1 & 307,373 & \ding{51} & \ding{51} & ~ & ~ & ~ & ~ & ~ & ~ & ~ & ~ & \ding{51} \\ 
2WikiMultiHopQA~\cite{DBLP:conf/coling/HoNSA20} & EN & 1 & 192,606 & \ding{51} & \ding{51} & ~ & ~ & ~ & ~ & ~ & ~ & ~ & ~ & \ding{51} \\ 
VisualMRC~\cite{DBLP:conf/aaai/TanakaNY21} & EN & 1 & 10,234 & \ding{51} & \ding{51} & \ding{51} & \ding{51} & ~ & ~ & ~ & ~ & ~ & ~ & \ding{51} \\ 
WebSRC~\cite{DBLP:conf/emnlp/ChenZCJZLX021} & EN & 1 & 6,400 & \ding{51} & \ding{51} & \ding{51} & \ding{51} & ~ & ~ & ~ & ~ & ~ & ~ & \ding{51} \\ 
MultiModalQA~\cite{talmor2021multimodalqa} & EN & 1 & 57,713 & \ding{51} & \ding{51} & \ding{51} & ~ & ~ & ~ & ~ & ~ & ~ & ~ & \ding{51} \\ 
DocVQA~\cite{DBLP:conf/wacv/MathewKJ21} & EN & 1 & 12,767 & \ding{51} & \ding{51} & \ding{51} & \ding{51} & ~ & ~ & ~ & ~ & ~ & ~ & \ding{51} \\ 
InforgraphicsVQA~\cite{DBLP:conf/wacv/MathewBTKVJ22} & EN & 1 & 5,485 & \ding{51} & \ding{51} & \ding{51} & \ding{51} & ~ & ~ & ~ & ~ & ~ & ~ & \ding{51} \\ 
MP-DocVQA~\cite{DBLP:journals/corr/abs-2212-05935} & EN & 1 & 5,928 & \ding{51} & \ding{51} & \ding{51} & \ding{51} & ~ & ~ & ~ & ~ & ~ & ~ & \ding{51} \\ 
DuReader$_{vis}$~\cite{DBLP:conf/acl/QiLLLZS0WL22} & CN & 1 & 158,000 & \ding{51} & \ding{51} & \ding{51} & \ding{51} & ~ & ~ & ~ & ~ & ~ & ~ & \ding{51} \\ 
DUDE~\cite{DBLP:conf/iccv/LandeghemPTJBBC23} & EN & 1 & 5,019 & \ding{51} & \ding{51} & \ding{51} & \ding{51} & ~ & ~ & ~ & ~ & ~ & ~ & \ding{51} \\ 
SlideVQA~\cite{DBLP:conf/aaai/TanakaNNHSS23} & EN & 1 & 2,619 & \ding{51} & \ding{51} & \ding{51} & \ding{51} & ~ & ~ & ~ & ~ & ~ & ~ & \ding{51} \\ 
xPQA~\cite{DBLP:journals/corr/abs-2305-09249} & Multi & 12 & 18,000 & \ding{51} & \ding{51} & ~ & ~ & ~ & ~ & ~ & ~ & ~ & ~ & \ding{51} \\ 
TAT-DQA~\cite{zhu2024doc2soargraph} & EN & 1 & 2,758 & \ding{51} & \ding{51} & \ding{51} & \ding{51} & ~ & ~ & ~ & ~ & ~ & ~ & \ding{51} \\ 
MMLongbench-Doc~\cite{DBLP:conf/nips/MaZC0JLLLMDZP0W24} & EN & 1 & 135 & \ding{51} & \ding{51} & \ding{51} & \ding{51} & ~ & ~ & ~ & ~ & ~ & ~ & \ding{51} \\ 
M-LongDoc~\cite{DBLP:journals/corr/abs-2411-06176} & EN & 1 & 851 & \ding{51} & \ding{51} & \ding{51} & \ding{51} & ~ & ~ & ~ & ~ & ~ & ~ & \ding{51} \\ 
VisDoMBench~\cite{DBLP:journals/corr/abs-2412-10704} & EN & 1 & 1,277 & \ding{51} & \ding{51} & \ding{51} & ~ & ~ & ~ & ~ & ~ & ~ & ~ & \ding{51} \\ 
ViDoRe~\cite{DBLP:journals/corr/abs-2407-01449} & Multi & 2 & 8,310 & \ding{51} & \ding{51} & \ding{51} & ~ & ~ & ~ & ~ & ~ & ~ & ~ & \ding{51} \\ 
X-AlpacaEva~\cite{DBLP:conf/acl/ZhangLF0J0B24} & Multi & 5 & 4,025 & \ding{51} & \ding{51} & ~ & ~ & ~ & ~ & ~ & ~ & ~ & ~ & \ding{51} \\ 
CHIC~\cite{DBLP:conf/icdar/MahamoudCJDO24} & Multi & 5 & 774 & \ding{56} & \ding{51} & \ding{51} & \ding{51} & ~ & ~ & ~ & ~ & ~ & ~ & \ding{51} \\
Belebele~\cite{DBLP:conf/acl/BandarkarLMASHG24} & Multi & 122 & 109,800 & \ding{51} & \ding{51} & ~ & ~ & ~ & ~ & ~ & ~ & ~ & ~ & \ding{51} \\ 
mOSCAR~\cite{DBLP:journals/corr/abs-2406-08707} & Multi & 163 & 314,827,611 & \ding{51} & \ding{51} & \ding{51} & ~ & ~ & ~ & ~ & ~ & ~ & ~ & \ding{51} \\ 
\bottomrule
\end{tabular}
\end{adjustbox}
\caption{Benchmark datasets for DAI tasks. Datasets are grouped by task, followed by publication year and language. }
\label{table:benchmark}
\end{table*}

\section{Multimodal Document AI (DAI)}
\label{sec_4:multimodal}
DAI has evolved significantly with the advent of multimodal AI, incorporating various information sources to comprehensively understand documents. 

\textbf{\textit{Textual modality} }
remains the primary information for most documents, whether extracted through OCR from scanned images or directly accessed from digital formats like PDFs. Early approaches treat text as a logical block sequence \cite{DBLP:conf/aaai/NiyogiS86}, forming the foundation for diverse document processing tasks. 
\textbf{\textit{Visual modalities}} contain rich signals, including formulae, figures, handwriting, and region designs ~\cite{DBLP:conf/iccvw/VianaO17,DBLP:conf/icdar/YiGLZLJ17}. These elements are essential for tasks like logo identification, signature verification, and analysis of non-textual layout components. Modern vision-language approaches have revolutionized document processing by integrating image features alongside text~\cite{DBLP:conf/mm/HuangL0LW22}, enabling end-to-end processing that transcends the limitations of traditional OCR-based methods. 
Moreover, \textbf{\textit{layout modalities}} focus on the spatial arrangement of document content, including bounding box coordinates, column structures, and region segmentation. This information is crucial for understanding relationships between document segments, such as key-value pairs within form regions. Advanced approaches either embed spatial information directly into the token sequence~\cite{DBLP:journals/corr/abs-2306-00526, DBLP:conf/iccv/HeWHLLXS23, DBLP:conf/acl/PerotKLSSBWWMZL24} or utilize separate encoders for bounding boxes~\cite{DBLP:journals/corr/abs-2407-01976,DBLP:conf/acl/0005RSMBKPNL24,DBLP:conf/cvpr/LuoSZZYY24}, enabling models jointly processing textual and spatial information. 
Additionally, some documents also take other types of modalities into consideration, \eg, tables~\cite{herzig2020tapas} and charts~\cite{luo2021chartocr}. 

Building on these multimodal capabilities, the next sections explore understanding and generation tasks in DAI, highlighting how LLMs are reshaping multimodal DAI to meet real-world needs.

\subsection{Multimodal Understanding in DAI}
\label{sec_4.1:multimodal_understanding}
Multimodal understanding in DAI involves \textit{analyzing}, \textit{extracting}, and \textit{classifying} document information by combining diverse modalities~\cite{srihari1986document, taylor1994integrated}. Leveraging LLMs, this field has evolved through two primary methods: \textit{prompt/instruction-based} and \textit{unified encoding} approaches, each offering unique advantages in fusing textual, visual, and layout modalities.

\subsubsection{Document Layout Analysis (DLA)}
DLA focuses on detecting and classifying structural elements within documents, such as text blocks, headers, as well as understanding their relationships. Beyond being a standalone task, DLA also serves as an auxiliary training objective to support downstream applications like KIE and Document QA~\cite{DBLP:conf/acl/YihCHG15}. 
Recent work such as~\citet{DBLP:conf/emnlp/FanWM024} explores symbolic self-training strategies to extract entity-level relationships within documents.
While large language models (LLMs) such as GPT-3.5, GPT-4o\footnote{https://chat.openai.com/chat}, and LLaMA2~\cite{DBLP:journals/corr/abs-2307-09288} have demonstrated strong general reasoning capabilities, their performance on DLA remains limited compared to specialized models like RoBERTa-based encoders~\cite{DBLP:journals/corr/abs-1907-11692}. 
Furthermore, methods such as in-context learning (ICL) and naive fine-tuning have shown limited success in improving LLMs' ability to capture document hierarchy and layout semantics~\cite{DBLP:conf/emnlp/XingCGSYBZY24,DBLP:conf/cvpr/LuoCZY23}.

To bridge this gap, two major lines of research have emerged for adapting LLMs to layout-aware document understanding: \textit{(1) Prompt/Instruction-based approaches} aim to inject layout modalities into LLMs through carefully designed prompts. For instance, layout-aware Chain-of-Thought (CoT) prompting has shown promising results in enhancing spatial reasoning during both pretraining and fine-tuning~\cite{DBLP:journals/corr/abs-2408-15045}. 
Other works~\cite{DBLP:conf/icdar/LamottWUSKO24, DBLP:conf/cvpr/LuoSZZYY24} demonstrate that layout-aware prompting can improve performance without requiring additional fine-tuning or structural supervision. \textit{(2) Unified encoding approaches} integrate text, spatial coordinates, and visual features into a shared representation. These models, such as LayoutLLM~\cite{DBLP:conf/cvpr/LuoSZZYY24} and DocLayLLM~\cite{DBLP:journals/corr/abs-2408-15045}, often adopt encoders like LayoutLMv3~\cite{DBLP:conf/mm/HuangL0LW22} and are trained with document segmentation or layout classification objectives. 
Inspired by human reading behavior,~\citet{DBLP:conf/emnlp/NguyenSSP21} propose to selectively attend to informative document regions. Meanwhile,~\citet{DBLP:conf/acl/0005RSMBKPNL24} incorporate disentangled spatial attention and infilling pretraining to model layout-aware field interactions.

\subsubsection{Key Information Extraction (KIE)}
KIE focuses on identifying and extracting specific elements from documents, such as form fields and key-value pairs.  
\textit{(1) Prompt-based approaches} have explored various methods of incorporating spatial information within prompts ~\cite{DBLP:conf/icdar/LamottWUSKO24,DBLP:conf/iccv/HeWHLLXS23} besides textual information, including bounding boxes, geometric formats and HTML markup. Similarly,~\citet{DBLP:conf/acl/PerotKLSSBWWMZL24} propose to leverage horizontal and vertical coordinates to capture spatial information. Additionally, LayTextLLM \cite{lu2024bounding} introduces "box tokens" to represent bounding boxes, aligning them with text using LoRA~\cite{DBLP:conf/iclr/HuSWALWWC22} for improved layout parsing. However, coordinate-as-token approaches significantly increase token length, leading to higher computational costs. To enhance spatial understanding, few-shot demonstrations~\cite{DBLP:conf/iccv/HeWHLLXS23} or fine-tuning~\cite{DBLP:conf/acl/PerotKLSSBWWMZL24} are often required, further exacerbating sequence length and GPU constraints. 
\textit{(2) Unified Encoding approach} is an efficient way with encoding multiple modalities into hidden vectors, allowing robust KIE from complex layout patterns~\cite{DBLP:conf/aaai/AppalarajuTDSZM24,DBLP:conf/acl/0005RSMBKPNL24}. For example, InstructDoc~\cite{DBLP:conf/aaai/TanakaINSS24} features an extra “Document-former” encoder that fuses OCR tokens with learnable tokens before forwarding them to the main LLM, enhancing field detection and entity extraction. Additionally, visually guided generative text-layout pretraining offers a hierarchical approach that functions as a native OCR model while simultaneously modeling text and layout~\cite{DBLP:conf/naacl/MaoBHS00W24}. Meanwhile, DoCo\cite{DBLP:conf/cvpr/LiWJGGCLJS24} utilizes contrastive learning to align fine-grained document-object features with holistic visual representations, addressing the challenge of feature collapse in text-rich scenarios.

\subsubsection{Document Classification (DC)} 
DC refers to the task of identifying the category, type, or domain of a document (\eg, invoice vs. resume) by leveraging textual, visual, and layout modalities.
\textit{(1) Prompt/Instruction-based approaches} utilize LLMs to perform document classification by framing the task as a natural language understanding problem. These methods allow zero-shot or few-shot generalization to unseen document types through instruction tuning or in-context exemplars. For example, LLMs have been shown effective in capturing document semantics and type-specific features when guided by descriptive prompts~\cite{DBLP:journals/corr/abs-2310-05690, DBLP:conf/iciis/HewapathiranaSA23}. However, their reliance on textual signals alone can limit performance in documents where layout or visual features are critical. \textit{(2) Unified Encoding approaches} integrate multiple modalities (\eg, OCR-based text, layout, and images) into a unified transformer model~\cite{DBLP:conf/acl/XuXL0WWLFZCZZ20, DBLP:conf/mm/HuangL0LW22}. These approaches excel in tasks where document layout (\eg, form structures) is a strong signal for classification~\cite{DBLP:conf/acl/0005RSMBKPNL24}, improving classification accuracy and efficiency. In this context,~\citet{DBLP:conf/icdar/PowalskiBJDPP21} employs a decoder capable of integrating textual, visual, and layout-based information for document understanding, achieving SoTA performance on classification and retrieval tasks while simplifying the end-to-end processing pipeline. And LayoutXLM~\cite{DBLP:journals/corr/abs-2104-08836} integrates text, layout, and image modalities to classify visually-enriched documents across different languages, demonstrating significant improvements in multilingual document contexts. By unifying multiple document-related modalities, the dependency on external OCR engines is mitigated. 
These OCR-free approaches can further reduce computational overhead and text-based bottlenecks~\cite{DBLP:conf/eccv/KimHYNPYHYHP22} by directly processing document images using a Transformer model. 
By leveraging the strengths of prompt-based, unified encoding, these models are pushing the boundaries of document analysis and classification capabilities. 

\subsection{Multimodal Generation in DAI}
\label{sec_4.2:multimodal_generation}
Multimodal document generation refers to the creation of new information based on the document input provided, either in the form of \textit{question-answers}, \textit{summaries}, or \textit{document elements}.  
With the advent of modern LLM-based techniques, they can be used as black-box tools for information generation via prompting to integrate OCR; or other methods as white-box models, embedding multimodal representations directly into the more controllable generation process.

\subsubsection{Document Question Answering (QA)} 
Document QA focuses on answering questions based on documents, involving complex table lookups, referencing figures, or extracting text from specific regions. 
Leveraging the efficacy of \textit{(1) Prompt/instruction-based approaches}, mPLUG-DocOwl~\citet{ye2023mplug} extends a vision-language LLM with document-specific instructions, utilizing strong image-text alignment learned during pretraining. It enables QA on scanned documents without explicit OCR. 
While ~\citet{DBLP:journals/corr/abs-2306-00526} demonstrates that LLMs can effectively capture layout information through manufactured prompts using spaces and line breaks, highlighting the potential of simple yet effective prompt engineering techniques.
Similarly, LATIN~\cite{DBLP:journals/corr/abs-2306-00526} preserves layout cues by inserting strategic spaces and line breaks in text prompts, significantly improving few-shot QA performance. 
Moreover, \textit{(2) Unified Encoding approaches} integrate multiple modalities into a single and cohesive framework, supporting end-to-end QA~\cite{DBLP:conf/wacv/MathewKJ21,DBLP:journals/corr/abs-2305-18565} without separate structure-parsing steps. \citet{DBLP:conf/emnlp/KimLKJPKYKLP23} employs a BART-style~\cite{DBLP:conf/acl/LewisLGGMLSZ20} encoder to process text and bounding box coordinates, coupled with an LLM decoder for answer generation. \citet{DBLP:conf/cvpr/ZhuLSGX0Z25} represents layout information as a single token and embeds the token through a layout projector module with a novel positional encoding scheme. In contrast, UDOP~\cite{DBLP:conf/cvpr/TangYWFLZ0ZB23} unifies vision, text, and layout features in a single Transformer, utilizing a generative mask-prediction framework to infill coherent answers conditioned on both textual and visual cues. Building on DocLLM~\cite{DBLP:conf/acl/0005RSMBKPNL24}, layout-aware decoding strategies integrate spatial and visual information directly into the generation process, enabling models to locate and interpret relevant regions even in irregular or complex document layouts and thereby improve answer accuracy.

\subsubsection{Document Summarization (DS)} 
DS aims to create concise overviews while preserving essential content, where recent advancements have explored various pathways. 
\textit{Visually-Enriched approaches} leverage large-scale document-level representations \cite{DBLP:conf/acl/ZhangWZ19, DBLP:conf/naacl/0001SRJ22} to enhance summarization performance, capturing comprehensive semantic information. Models like PaLI-X \cite{DBLP:journals/corr/abs-2305-18565} demonstrate versatility by encoding entire pages, including text and images, to generate structured summaries. While image-text alignment techniques strengthen cross-modal coherence through pseudo image captions \cite{DBLP:conf/acl/Jiang00SZ23}, effectively integrating visual and textual information. Moreover, \textit{Structure-Aware approaches} incorporate additional modalities beyond text-image integration to enrich document summaries, where \citet{DBLP:conf/emnlp/00020YW22} integrates tabular data into financial and business reports, ensuring table-aware insights are effectively reflected in summaries. And layout-aware summarization utilizes document layout analysis to capture both physical and logical structures, maintaining the integrity of the original document's information flow. 

Besides the aforementioned approaches, strategies have been developed to address challenges while handling \textit{longer or multiple documents}. Techniques focus on distilling key information ~\cite{DBLP:conf/acl/XiaoBCC22, DBLP:conf/ecai/LiuLYJLZLMH24}, reducing redundancy across multiple resources.~\citet{DBLP:journals/corr/abs-2306-15595} have employed position interpolation, extending context windows to maintain coherence across large volumes of text within long documents. A monotone-submodular content selection approach has emerged to prioritize essential events across multiple sources \cite{DBLP:journals/corr/abs-2310-03414}. These developments reflect the ongoing evolution of producing more coherent, concise, and comprehensive overviews of complex materials.

\subsubsection{Document Content Generation (DCG)} 
DCG focuses on automatically generating structured document layouts and textual content, emphasizing design and coherence in document composition. \textit{Layout synthesis} seeks to dynamically structure documents based on input semantics. Advanced from heuristic rules to learning implicit relationships between textual and visual elements, deep generative models for content-aware graphic design modeling \cite{DBLP:journals/tog/ZhengQCL19} generate adaptable layouts aligned with content themes. 
LLM-based autoregressive document modeling goes beyond layout-specific generation, incorporating sequential dependencies between textual and structural elements. 
In this context, \citet{DBLP:journals/corr/abs-2406-08354} jointly models document structure and content, synthesizing cohesive documents without relying solely on visual components. \textit{Multimodal document synthesis} aims to generate textual, visual, and structural modalities with richer document representations. 
Methods like UDOP~\cite{DBLP:conf/cvpr/TangYWFLZ0ZB23} leverage large-scale pretraining to unify vision, text, and layout, enabling comprehensive document understanding and generation across diverse domains.  Similarly, StrucTexTv2 \cite{DBLP:conf/iclr/YuLZZGQYHD023} introduces masked visual-textual pretraining, improving document structure modeling while eliminating OCR dependencies, achieving robust document processing. 
Meanwhile, some methods tailored for specialized document elements, such as ChartLlama \cite{DBLP:journals/corr/abs-2311-16483}, focus on improving chart understanding and generation by leveraging multimodal instruction tuning. 
\section{Multilingual Document AI}
\label{sec_3:multilingual}
There are over 7,000 spoken languages worldwide~\cite{DBLP:conf/acl/JoshiSBBC20} and a significant portion of online content is written in languages other than English~\cite{DBLP:journals/corr/abs-2104-08836}. 
This poses challenges to support multilingual DAI. Models trained on monolingual data unsurprisingly struggle with multilingual tasks due to their limited capacity to capture cross-linguistic nuances and cultural intricacies.
In contrast, LLMs, trained on vast multilingual data, exhibit remarkable capabilities in DAI tasks~\cite{DBLP:conf/acl/BandarkarLMASHG24, DBLP:conf/naacl/0001SRJ22, DBLP:journals/corr/abs-2211-05100, DBLP:journals/corr/abs-2404-07613}, though challenges still remain.

\subsection{Multilingual Prompt-Based Approaches}
Many studies examine LLMs’ zero-shot prompting capabilities across languages, revealing that they often outperform fine-tuned small models in KIE and DSA. However, their superiority is not consistently observed in other tasks~\cite{DBLP:journals/corr/abs-2309-09800, DBLP:conf/eval4nlp/BhatV23}. Factors such as prompt quality and supplementary instructions may limit performance. \citet{DBLP:conf/emnlp/LaiNVMDBN23} find ChatGPT underperforms fine-tuned models on multilingual tasks but excels when prompted in English, a trend also seen in~\cite{DBLP:journals/corr/abs-2409-04512, 2305.13582}. Cross-lingual thought prompts ~\cite{DBLP:conf/emnlp/HuangTZZSXW23} and self-translate ~\cite{DBLP:conf/naacl/EtxanizASLA24} further explore translation instructions into English.

ICL with non-English examples boosts KIE and DC performance~\cite{DBLP:journals/corr/abs-2309-09800}, though the effectiveness is inconsistent~\cite{DBLP:conf/emnlp/EnglanderSPPKG24}. To address this, strategies such as incorrect examples~\cite{DBLP:conf/emnlp/MoLYWZWL24}, multilingual semantic similarity~\cite{DBLP:conf/acl/TanwarDB023}, and word-level code-switching~\cite{shankar2024context} have been employed. Other prompt-based approaches include soft prompt tuning for cross-lingual relation extraction~\cite{DBLP:conf/ijcnn/HsuZDWWLLH23}, automated prompt construction from relation triples~\cite{DBLP:conf/emnlp/ChenHH22}, and multi-turn QA for zero-shot KIE~\cite{wei2024chatie}. Despite these efforts, challenges persist in multilingual DAI, such as prompt sensitivity and representation gaps, which are constrained by LLMs' inherent multilingual capabilities.

\subsection{Multilingual Training Strategies}
English-centric LLMs often struggle with capturing cultural and linguistic nuances, leading to limitations in DAI~\cite{DBLP:conf/acl/HershcovichFLLA22, DBLP:journals/jdiq/NavigliCR23}. To address this, PersianLLaMA~\cite{DBLP:journals/corr/abs-2312-15713} and JASMINE~\cite{DBLP:conf/emnlp/NagoudiAEIK23} incorporate curated document collections for better alignment, while ArabianGPT~\cite{DBLP:journals/corr/abs-2402-15313} improves Arabic morphological processing through an optimized tokenizer. Additionally, preference-tuning techniques including reinforcement learning with AI feedback (RLAIF)~\cite{DBLP:conf/icml/0001PMMFLBHCRP24} and direct preference optimization (DPO)~\cite{DBLP:conf/nips/RafailovSMMEF23} have been applied to align on cultural and linguistic preferences~\cite{DBLP:conf/naacl/HuangYZSCSCAAHL24, DBLP:journals/corr/abs-2406-16316}. \citet{DBLP:conf/acl/LaiM024} propose to translate English instructions and inputs into multiple target languages and align with human preference, showing promising performance on knowledge-intensive DAI tasks.

On the other hand, training on large-scale multilingual documents may introduce language interference, \ie, the curse of multilinguality~\cite{DBLP:conf/emnlp/ChangAT024}. To address this, multi-stage training paradigms are adopted, selectively fine-tuning different model components at various stages~\cite{DBLP:journals/eswa/LiWCYBYLGYAL24, DBLP:conf/cecct/WangYLCLMBL23}, providing more accurate KIE capabilities. Other approaches~\cite{DBLP:journals/corr/abs-2407-19672, DBLP:conf/naacl/KojimaOIYM24} apply language-specific neurons~\cite{DBLP:journals/corr/abs-2402-18815} to train with each monolingual data while preserving high-resource languages performance. Also, parameter-efficient techniques such as LoRA and adapters~\cite{DBLP:journals/eswa/LiWCYBYLGYAL24, DBLP:conf/naacl/WhitehouseHBDLL24, DBLP:journals/corr/abs-2402-18913} enable language-sensitive adaptation.

Another core challenge in multilingual DAI is achieving effective cross-lingual alignment. 
\citet{DBLP:conf/aaai/NooralahzadehS23} improves alignment through similarity-based loss and synthetic code-mixing.
PLUG~\cite{DBLP:conf/acl/ZhangLF0J0B24} enhances linguistic structure alignment via pivot-language instructions. Contrastive learning, particularly effective for retrieval tasks~\cite{DBLP:conf/emnlp/WangWN22, DBLP:conf/eacl/TanHSK23}, improves alignment by clustering positive examples while separating negatives. Another approach leverages external parallel data, aligning text at the sentence~\cite{DBLP:journals/corr/abs-2308-04948, DBLP:conf/emnlp/HeffernanCS22} or word level~\cite{DBLP:conf/starsem/ZhaoEBA21, DBLP:conf/acl/Chi0ZHMHW20}.
Additionally, data augmentation strengthens alignment by generating synthetic data via translation~\cite{DBLP:journals/corr/abs-2411-16300} or self-distillation~\cite{DBLP:conf/acl/ZhangWLWWL0024}. \citet{DBLP:journals/corr/abs-2410-14815} propose to use balanced synthetic corpora to continue pretraining LLMs.

These advancements underscore the potential of multilingual LLMs in DAI, though challenges in representation and effective cross-lingual generalization remain open research directions.
\section{Retrieval-Augmented Paradigm}
\label{rag}

The retrieval-augmented paradigm has been widely studied in LLM research and industry. Many studies~\cite{DBLP:journals/corr/abs-2312-10997, DBLP:journals/tacl/RamLDMSLS23, DBLP:conf/kdd/FanDNWLYCL24, DBLP:journals/corr/abs-2406-15187} have shown that retrieving reliable external knowledge can mitigate challenges associated with outdated training data and limited domain expertise. Documents, encompassing multiple modalities like text, tables, and images, can be essentially integrated into this paradigm to support DAI~\cite{DBLP:conf/emnlp/ZhaoCWJLQDGLLJ23, DBLP:conf/iclr/SomanR24}. For example, in business document applications, retrieval-augmented generation (RAG) is integrated into a document intelligence service, leveraging semantic chunking and layout information~\cite{azure_ai_document}. Similarly, it has been utilized to provide document grounding capabilities, allowing users to leverage extra documents for QA~\cite{sap_joule}.

\subsection{Text-Based Retrieval Augmentation}
\emph{Text}, as the core component of documents, holds a wealth of information. Text-based retrieval augmentation can provide LLMs with precise and reliable contextual information, supporting them in understanding and generating tasks. \citet{DBLP:journals/corr/abs-2002-08909} introduce a pretrained retrieval-augmented language model (REALM). RAG is proposed around the same time, combining a pretrained model with a non-parametric memory storing a dense vector index of the external database~\cite{DBLP:conf/nips/LewisPPPKGKLYR020}. Subsequently, many studies have been conducted to enhance the performance and robustness of document-related tasks~\cite{DBLP:conf/icml/BorgeaudMHCRM0L22, DBLP:journals/jmlr/IzacardLLHPSDJRG23, DBLP:journals/tacl/RamLDMSLS23, DBLP:conf/emnlp/00110PCML0024, DBLP:journals/corr/abs-2401-15884, DBLP:journals/corr/abs-2403-10131}.

For \emph{KIE}, specifically event argument extraction, \citet{DBLP:conf/naacl/LiJH21} propose a conditional model based on a predefined template, designed to handle missing arguments and enable cross-sentence inferences. Retrieve-and-Sample~\cite{DBLP:conf/acl/RenCG0M023} is proposed to address the challenge of similar inputs leading to inconsistent outputs, which utilizes a hybrid RAG that samples pseudo demonstrations. DS and QA also benefit from RAG, \eg, ~\citet{DBLP:journals/corr/abs-2404-16130} propose Graph RAG for query-focused summarization and DR-RAG~\cite{DBLP:journals/corr/abs-2406-07348} is introduced for QA by evaluating the dynamic relevance between query and documents.

For \emph{long-context} DAI tasks, \citet{DBLP:conf/iclr/0008PWM0LSBSC24} find that using RAG can achieve comparable performance to fine-tuned LLM while taking much less computation. To improve long-context document retrieval, RAPTOR~\cite{DBLP:conf/iclr/SarthiATKGM24} is introduced to integrate information from a tree structure with differing levels of summarization to improve holistic document understanding. \citet{DBLP:journals/corr/abs-2405-18414} introduce a document-graph-based re-ranker to improve RAG. Besides, to evaluate the performance of long-document retrieval, \citet{DBLP:conf/emnlp/LabanFXW24} introduce a synthetic testbed derived from summarization tasks and demonstrated that it plays a central role in assessing RAG effectiveness in long-context scenarios. Despite these efforts, long-context document retrieval remains a challenge, particularly in understanding how retrieval augmentation affects LLMs and how retrieved documents can be effectively incorporated into LLMs~\cite{DBLP:journals/corr/abs-2310-12150}.

\subsection{Multimodal Retrieval Augmentation}
Multimodal retrieval augmentation goes beyond using only text information, incorporating images, charts, and tables as retrieval targets~\cite{DBLP:conf/emnlp/ZhaoCWJLQDGLLJ23}. By interpreting and leveraging varied data modalities, it aims to boost the performance of LLMs in downstream tasks~\cite{DBLP:conf/kdd/FanDNWLYCL24}.

\emph{Heuristic approaches} use different parsing tools for various modalities to obtain corresponding embeddings from given documents. RA-VQA~\cite{DBLP:journals/corr/abs-2210-03809} integrate OCR, object detection, and image captioning to convert target images into textual data for subsequent retrieval. Meanwhile, tables in documents often appear in a semi-structured format, which requires table detection and parsing techniques~\cite{DBLP:journals/pr/MaLSH23, DBLP:journals/corr/abs-2401-12599}. For structured tabular data, T-RAG~\cite{DBLP:journals/corr/abs-2203-16714} utilizes vector index jointly with BART~\cite{DBLP:conf/acl/LewisLGGMLSZ20} to address QA. By leveraging table-to-table relevance and knowledge graphs to represent relationships between structured data, \citet{DBLP:conf/acl/ChenZR24} and \citet{sepasdar2024enhancing} enhance structured document retrieval.

Recently, advancements in vision-language models (VLMs)~\cite{DBLP:journals/corr/abs-2303-08774, DBLP:conf/nips/LiuLWL23a, DBLP:journals/corr/abs-2409-12191} have enabled \emph{end-to-end multimodal retrieval}, which reduces information loss and minimizes error accumulation. DSE~\cite{DBLP:journals/corr/abs-2406-11251} is introduced to capture all information within a document by using screenshots and leveraging VLMs to encode these into retrieval representations. Similarly, ColPali~\cite{DBLP:journals/corr/abs-2407-01449} utilizes ColBERT~\cite{DBLP:conf/sigir/KhattabZ20} representations to index page-level documents, while VisRAG~\cite{yu2024visrag}, built on MiniCPM-V~\cite{DBLP:journals/corr/abs-2408-01800}, supports page concatenation to improve retrieval performance. For visual document QA, VisDoMRAG~\cite{DBLP:journals/corr/abs-2412-10704} incorporates evidence curation and CoT to facilitate textual and visual RAGs, where the outputs are aligned to ensure consistency. Despite the significant development in multimodal retrieval, multimodal foundation models, effective embedding and representation methods, and comprehensive evaluation approaches remain open research directions.

\section{Future Directions}
\label{agent}
AI agents are intelligent systems that exhibit a range of cognitive capabilities, including perception, learning, memory, planning, decision-making, action execution, and collaboration with other agents or humans~\cite{DBLP:journals/corr/abs-2405-06643,DBLP:journals/corr/abs-2403-00833}. Recent research has integrated reasoning and action within language models to tackle complex decision-making tasks~\cite{DBLP:conf/iclr/YaoZYDSN023}, while advancements in learning from mistakes~\cite{DBLP:conf/nips/ShinnCGNY23} have further enhanced the performance of LLM agents. These developments have motivated applications across various fields, such as healthcare~\cite{DBLP:journals/corr/abs-2407-02483,DBLP:journals/npjdm/MehandruMASBA24} software~\cite{DBLP:conf/acl/ZhangLLSJ24,DBLP:conf/acl/QianDLLXWC0CCL024,tang2024collaborative}, and finance~\cite{DBLP:journals/corr/abs-2407-09546,DBLP:journals/corr/abs-2405-14767}, demonstrating the transformative potential of AI agents. The rise of multilingual and multimodal LLMs~\cite{DBLP:journals/corr/abs-2303-08774} has positioned AI agents for document processing as a promising research area, offering numerous opportunities for innovation in handling complex document-related tasks. 

Building on these insights, Document LLM agents (DocAgents) hold great potential for revolutionizing document processing. We define an autonomous DocAgent as an intelligent system designed to proficiently manage document understanding and generation tasks, achieving a level of expertise comparable to human specialists.

\subsection{Collaborative DocAgent Framework}
Documents typically combine rich textual representations, visual elements, and dynamic layouts~\cite{DBLP:journals/corr/abs-2408-01287, DBLP:conf/emnlp/ZhangTZYCZCGZZG24}. Managing these modalities often requires a specialized multi-agent framework, where effectively navigating these diverse modalities necessitates domain-expert agents. In this context, LLM-based multi-agent systems can collaborate by leveraging domain-specific knowledge, external tools, system feedback, and human inputs~\cite{DBLP:journals/corr/abs-2409-13753}. 

While existing frameworks have shown promising results in document simplification~\cite{DBLP:conf/emnlp/MoH24}, they still encounter significant challenges in comprehending visually rich documents. Overcoming these obstacles requires integrating collaborative image-text retrieval capabilities and enhanced representations for complex document layouts, enabling a more comprehensive and nuanced approach to address complicated document-related tasks. Recent research indicates that even advanced multimodal LLMs struggle with structured data tasks, such as table representations~\cite{DBLP:conf/acl/DengSHS0M0M24}. To advance DocAgent frameworks, future research endeavors should focus on:
\textit{(1) Enhanced Reasoning and Generalization:} Developing systems that produce documents tailored to specific input constraints, such as layout templates and natural language instructions. \textit{(2) Complex Layout Handling:} Improving capabilities to manage dynamic, multimodal information within diverse document layouts. \textit{(3) Efficient Information Retrieval:} Enabling global and local information searches with open-world interactions to capture relevant semantic nuances. 
Addressing these long-term challenges is key to developing reliable, explainable DocAgents for diverse document processing tasks.

\subsection{DocAgent Foundation Model}
While LLMs have emerged as general-purpose foundation models~\cite{DBLP:journals/corr/abs-2108-07258}, they face two notable limitations:
\textit{(1)	Domain-Specific Knowledge:} LLMs trained on broad web data often underperform on specialized tasks~\cite{DBLP:conf/acl/XieAA24,DBLP:conf/acl/DengSHS0M0M24}. \textit{(2)	Cross-Modal Alignment:} Current methods that utilize noisy web-text pairs and frozen encoder-decoder architectures can lead to misalignment across different modalities~\cite{DBLP:conf/icml/0001LXH22,DBLP:conf/nips/AlayracDLMBHLMM22}. These limitations can hinder the model's ability to integrate and process information from documents. 

Recent efforts have introduced interactive foundation models that adapt to specific domains, such as robotics and healthcare~\cite{DBLP:journals/corr/abs-2402-05929}. To address document-specific challenges, future work should prioritize: \textit{(1) Cross-Modal Datasets:} Developing high-quality paired datasets that capture text, visual elements, and layout dynamics~\cite{DBLP:conf/emnlp/ZhangTZYCZCGZZG24,DBLP:conf/emnlp/XingCGSYBZY24,DBLP:conf/emnlp/LiHLC0LLH024}.
\textit{(2)	Template-Based Information Extraction:} Leveraging question templates to extract key information with greater precision~\cite{zmigrod-etal-2024-value}. \textit{(3)	Enhanced Alignment Techniques:} Incorporating domain-specific question-answer agents to verify and refine multimodal data, ensuring consistency and accuracy across different modalities.

Developing a domain-aware DocAgent foundation model, supported by continuously evolving datasets, paves the way for an end-to-end solution for complex document processing tasks. 
\section{Conclusions}
This survey provides a comprehensive review of recent advancements in Document AI (DAI), with a particular focus on LLM-based approaches. We have systematically outlined the evolving landscape of DAI, categorizing current tasks into two categories: understanding and generation. 
Our exploration reveals significant progress in integrating textual, layout, and visual modalities through prompt engineering or unified encoding strategies to handle DAI tasks. We also highlight the transformative impact of multilingual LLMs and retrieval-augmented methods on document processing capabilities, while also identifying remaining challenges in cross-lingual generalization, structural comprehension, and efficient multimodal learning. Last but not least, we discuss emerging trends, such as agent-based frameworks, as a promising path toward more robust and adaptable DAI systems.
\section*{Limitations}

While we have strived to provide a comprehensive overview of recent advancements in the field of Document AI, certain limitations are inevitable.

First, our selection of works primarily focuses on cutting-edge LLM-based methods published in major conferences such as ACL, EMNLP, NAACL, NeurIPS, ICLR, and preprint repositories like arXiv over the last three years due to space constraints.
While this focus reflects current trends, it may inadvertently lead to the exclusion of relevant contributions published in other venues or emerging close to or after the completion of this survey.

Second, while this survey emphasizes LLM-based approaches to align with recent developments in Document AI, non-LLM methods remain highly relevant and, in certain tasks and domains, may even be more effective. However, due to our focus, these approaches are not comprehensively covered, and their comparative advantages warrant further exploration.

Last but not least, the benchmarks discussed in this survey are among the most widely adopted in the Document AI community. While they provide a representative view of current evaluation practices, we acknowledge that other important benchmarks may not be fully captured. Future work could benefit from a broader investigation of evaluation metrics and benchmark datasets.

\bibliography{custom}

\appendix
\clearpage
\section{Evolution: Early Approaches}
\label{appx:evolution}
In this section, we review early multimodal and multilingual approaches for document-related understanding and generation tasks.

\subsection{Understanding Tasks}

\subsubsection{Multimodal}

Before the rise of LLMs, multimodal document understanding relied heavily on rule-based systems, template matching, and traditional machine learning. Key early models integrate text and layout information using feature engineering techniques.

\paragraph{Rule-based} systems rely on manually defined sets of rules and logic to interpret and process data. ~\citet{DBLP:conf/aaai/NiyogiS86} introduces a rule-based system that analyzes digitized document images. The system identifies various printed regions within the document and classifies them into logical “blocks” of information. It employs a goal-directed, top-down approach, utilizing a three-level rule hierarchy to guide its control strategy. 
~\citet{DBLP:journals/ijdar/KlinkK01} presents a rule-based system that combines layout and textual features to understand document structure. The system uses fuzzy logic to merge these features, allowing for adaptable rule bases that can be tailored to new domains. It processes documents in two stages: document analysis and document understanding. 
~\citet{DBLP:journals/corr/abs-2009-06611} proposes a knowledge-based method for assembling legal documents. It utilizes a machine-readable representation of legal professionals’ knowledge, comprising a rule base for formal legal norms and document templates for tacit knowledge. The system collects input data through an interactive interview, performs legal reasoning over this data, and generates the output document. It also creates an argument graph to explain the reasoning process, providing users with an interpretation of how input data and the rule base influence the output document.

\paragraph{Template Matching} is a technique where predefined patterns or “templates” are compared to input data (such as text or images) to identify similarities.
~\citet{dhakal2019one} introduces a novel one-shot template-matching algorithm designed to automatically capture data from business documents, aiming to minimize manual data entry. The algorithm utilizes a set of engineered visual and textual features, enabling it to be invariant to changes in position and value.~\citet{nguyen2024robust} presents a component-based template matching approach that leverages document layout analysis. This method involves sub-graph mining to extract significant sub-graphs from a training set, identifying recurrent patterns. A deep graph neural network is then employed to align these sub-graphs with segments in the query document, enhancing the robustness of the template matching algorithm.

\paragraph{Traditional Machine Learning} involves algorithms that learn patterns from data without being explicitly programmed with rules. In the visual domain, ~\citet{DBLP:conf/iccvw/VianaO17} and ~\citet{DBLP:conf/icdar/YiGLZLJ17} first introduce R-CNN~\cite{DBLP:conf/cvpr/GirshickDDM14} into DLA task, followed by studies~\cite{DBLP:journals/corr/abs-2201-11438, DBLP:conf/icdar/SahaMJ19, DBLP:conf/icip/YangH22, DBLP:conf/icdar/WangHH24} adopting more advanced detection and segmentation models. The progress also fuels research interest in applications for diverse document types and specific document components such as element segmentation for newspaper~\cite{DBLP:conf/icmla/AlmutairiA19} and historical documents~\cite{DBLP:conf/icfhr/OliveiraSK18}, keyword detection in Chinese documents~\cite{lin2021keyword} and graphical object detection~\cite{DBLP:conf/icdar/SahaMJ19} for mathematical equations and figures. To handle lengthy text, LSTM and transformer module are widely adopted~\cite{DBLP:conf/emnlp/WangX0SW21, DBLP:conf/icdm/RauschRG0F23, DBLP:conf/icdar/WangHH24}. 
More recent works design specific module to interact with modalities uniquely present in documents, such as layout and chart. 
Chargrid \cite{katti2018chargrid} is one of the first models to process documents as 2D grids, treating document content as both text and spatial layout.
TRIE \cite{zhang2020trie} combined text recognition with information extraction, leveraging end-to-end methods but still using handcrafted features for layout understanding. ~\citet{DBLP:conf/icpr/YuLQG020} and ~\citet{DBLP:conf/naacl/LiuGZZ19} treat documents as a graph and embed width and height of text segment as well as vertical and horizontal distance between two segments as graph embedding. With the advancement of the transformer-based model, coordinate encoding gradually becomes the default paradigm. LayoutLM~\cite{DBLP:conf/kdd/XuL0HW020} pre-trains BERT models with coordinate information and image embeddings integrated with text and positional embeddings. Follow-up works either incorporate additional pre-training tasks~\cite{DBLP:conf/acl/XuXL0WWLFZCZZ20,DBLP:conf/aaai/HongKJHNP22,DBLP:conf/mm/HuangL0LW22} and reading order~\cite{DBLP:conf/cvpr/GuMWLWG022} from layout information or inject 2D token position into transformer's attention bias~\cite{DBLP:conf/icdar/PowalskiBJDPP21,DBLP:conf/icdar/GarncarekPSTHTG21,DBLP:conf/cvpr/TangYWFLZ0ZB23}. DocMamba~\cite{DBLP:journals/corr/abs-2409-11887} explores Mamba~\cite{DBLP:journals/corr/abs-2312-00752} in DAI tasks, making use of layout segment and token coordinates to derive 1D text tokens and 2D position embeddings.

\subsubsection{Multilingual}

\paragraph{Language Model Pre-training} leverages advancements in transformer-based unsupervised pre-training on large-scale multilingual datasets, employing a training objective specifically designed to optimize performance on multilingual tasks. ~\citet{DBLP:conf/naacl/ChiDWYSWSMHZ21} pre-trained a cross-lingual model InfoXLM by maximizing the mutual information at multi-granularity level, including monolingual token-sequence, cross-lingual token-sequence and cross-lingual sequence-level. LayoutXLM \cite{DBLP:journals/corr/abs-2104-08836} combines the multimodal and multilingual pre-training techniques from LayoutLMv2 \cite{DBLP:conf/acl/XuXL0WWLFZCZZ20} and InfoXLM respectively. During the pre-training stage, aside from text-image alignment and matching, a novel multilingual masked visual-language modeling objective is proposed. Evaluation on language-specific finetuning, zero-shot transfer learning and multitask fine-tuning showed significant improvement over XLM-RoBERTa and InfoXLM. In the cross-lingual information retrieval (CLIR) task,~\citet{DBLP:conf/cikm/WangZZGW021} fine-tunes mBERT \cite{DBLP:conf/naacl/DevlinCLT19} with deep relevant matching objective on a home-made CLIR training data derived from parallel corpora, showing promising results on the retrieval of Lithuanian documents against short English queries.

\paragraph{Language-specific Training} The imbalance of data distribution in multilingual training often leads to poorer performance on Low-resource Languages (LRLs) than High-resource Languages(HRLs) \cite{DBLP:journals/corr/abs-2404-11553}. While the models are trained on a variety of languages, the performance in downstream tasks declines, a phenomenon known as the curse of multilinguality \cite{DBLP:conf/emnlp/ChangAT024}. ColBERT-XM \cite{DBLP:journals/corr/abs-2402-15059} applies language-specific adapters in pre-training. Adapters and embedding layers were frozen during monolingual fine-tuning to force learning in the shared weights from high-resourced languages. Low-resource languages can benefit from data-efficient training via updating their adapters only. Similar work \cite{DBLP:journals/corr/abs-2212-10448} on CLIR shows that models trained with monolingual data are more effective than fine-tuning the entire model.

\paragraph{Language-independent Approach} assumes features other than text could remain invariant across documents in different languages. LiLT~\cite{DBLP:conf/acl/WangJD22} separates text and layout features during the pre-training stage, enhancing layout structure learning using monolingual data. Multilingual fine-tuning further boosts cross-lingual capabilities while requiring less supervised data. Inspired by LiLT, ~\citet{shen2024ldp} further decouples visual features during pre-training. Since image contains language information, they adopt a text edit diffusion model to replace the original texts with fictional words.

\paragraph{Translation-aided Approach} ~\citet{DBLP:conf/acl/LiuDBJSM20} introduces a data augmentation framework via translation for low-resource named entity recognition (NER) task. The sentences are input into the translation system in two modified forms of the original source sentence. In the first form, entity words are replaced with corresponding entity tags and indexes. In the second form, brackets are used to include word span information alongside the entity words. After translation, the entity tags in the output are replaced with the translations extracted from the bracketed spans.

\subsection{Generation Tasks}

\subsubsection{Multimodal}

While effective for understanding document structure, \textbf{statistical methods} often fall short in generating new content, as they primarily rely on existing textual patterns. Instead, early works adopt the following approaches.

\paragraph{Heuristic-based Approaches} rely on manually designed rules or strategies to generate content. To generate summarization, these approaches analyze specific text features \cite{carbonell1998use}, such as keyword frequency \cite{luhn1958automatic} (using metrics like TF or TF-IDF to identify key sentences), sentence length, positional importance (prioritizing content at the beginning or end of a document), and title matching (selecting sentences with title-related keywords) \cite{edmundson1969new}. To generate answers, heuristic systems answer the questions from predefined patterns within a closed domain \cite{gupta2012survey}. Similarly, to generate document classification tags, feature engineering techniques like tokenization, stop-word removal, and stemming \cite{wang2005new} are used to identify linguistic patterns. 
While straightforward, these methods often lack the flexibility of more advanced approaches.

\paragraph{Optimization-based Approaches} aim to mathematically identify the “best” unseen content under specific constraints. 
To generate summarization, these methods optimize objectives such as maximizing information coverage \cite{takamura2009text} or minimizing redundancy \cite{hirao2013single} to achieve an ideal compression ratio. Some notable techniques include linear programming \cite{mcdonald2007study} for selecting sentence subsets that satisfy predefined criteria and sub-modular optimization \cite{lin2011class} to balance diversity and coverage. In question answering, optimization frameworks enhance the precision and recall of retrieved answers, while in document classification, they are used to refine feature selection and representation, enabling the effective handling of lengthy and complex documents.

\paragraph{Traditional Machine Learning} has advanced document generation through deep learning and pre-trained models. To generate summarization, supervised learning trains neural networks on document-summary pairs \cite{DBLP:conf/aaai/NallapatiZZ17, narayan2018ranking, see2017get}, using word embeddings like Word2Vec \cite{mikolov2013efficient} and GloVe \cite{pennington2014glove}. 
To generate answers against document-related questions, more modalities are taken into consideration, such as images and charts.
Datasets like OCR-VQA \cite{DBLP:conf/icdar/0001SSC19}, TextVQA\cite{singh2019towards}, and DocVQA\cite{DBLP:conf/wacv/MathewKJ21} focus on single-image QA, while MultiModalQA \cite{talmor2021multimodalqa} and MP-DocVQA \cite{DBLP:journals/corr/abs-2212-05935} emphasize reasoning across multiple pages or images. 
Generative models in table-based QA directly generate answers, such as Seq2Seq \cite{dong2016language,zhong2017seq2sql,wang2018robust} and graph-based approaches \cite{mueller2019answering}. Seq2Seq models focusing on non-database tables and free-form QA. Graph-based methods represent tables as graphs but are limited to table-only tasks.
To generate document tags for classification, machine learning pipelines \cite{han2024length} utilize tokenization, vectorization, and embedding to model high-dimensional text data. Neural networks, particularly transformers, excel at handling lengthy and multi-paragraph documents \cite{tuteja2023long}.

\subsubsection{Multilingual}

\paragraph{Pipeline Method} decomposes the original task into sub-tasks and involves machine translation either as the first step or the last step to avoid resource-intensive multilingual annotation. ~\citet{DBLP:journals/talip/LeuskiLZGOH03} first translates Hindi documents to English using a statistical machine translation system, after which important English sentences are selected to create a summary. ~\citet{DBLP:conf/sac/AraujoRPB16} compares twenty-one methods and two language-specific models on nine language-specific dataset. The finding suggests that simply translating the input text to English and leveraging English-based methods is better than language-specific methods. ~\citet{DBLP:conf/acl/DuhFN11} argues that cross-lingual errors are not caused by machine translation errors, and would occur even with a perfect translation model. A better cross-lingual algorithm is needed to mitigate the errors.

\paragraph{Pre-trained Model Approach}  The emergence of per-trained models has benefited almost every area of NLP, including multilingual document generation tasks. By simply fine-tuning on large-scale cross-lingual summarization datasets, it's able to outperform many multi-task models \cite{DBLP:conf/acl/LiangMZXCSZ22}. Similar work \cite{DBLP:journals/corr/abs-2309-15779} also shows the effectiveness of mBERT on cross-lingual QA tasks.

\paragraph{Cross-lingual Transfer} explores the language transfer capability of models with limited multilingual corpora. ~\citet{DBLP:journals/corr/abs-1907-06042} explores cross-lingual transfer learning for question answering by leveraging a source language with abundant annotations to improve performance in a target language with limited data. A GAN-based approach is proposed to incorporate a language discriminator to learn language-universal feature representations, and consequentially transfer knowledge from the source language. ~\citet{DBLP:conf/ijcnlp/XuZSZH20} proposes a mixed-lingual pre-training approach that leverages both cross-lingual tasks, such as translation, and monolingual tasks, like masked language modeling, to enhance cross-lingual summarization. The model effectively utilizes massive monolingual data to improve language modeling and achieves significant improvements in ROUGE~\cite{lin-2004-rouge} scores over state-of-the-art results on the NCLS dataset~\cite{DBLP:conf/emnlp/ZhuWWZZWZ19}.

\end{document}